\documentclass{article}
\usepackage{arxiv}
\usepackage{fancyhdr}
\usepackage[utf8]{inputenc} 
\usepackage[T1]{fontenc}    
\usepackage{hyperref}       
\usepackage{url}            
\usepackage{booktabs}       
\usepackage{amsfonts}       
\usepackage{nicefrac}       
\usepackage{microtype}      
\usepackage{lipsum}
\usepackage{graphicx}
\usepackage{enumitem}       
\graphicspath{ {./images/} }

\usepackage{array, tabularx, ragged2e}
\usepackage{array,multirow}
\newcolumntype{P}[1]{>{\raggedright\arraybackslash}p{#1}}

\title{A Practical-Driven Framework for Transitioning Drive-by-Wire to Autonomous Driving Systems: A Case Study with a Chrysler Pacifica Hybrid Vehicle\thanks{This paper has been accepted for publication by the 22nd International Conference on Automation Technology (Automation 2025). © 2025 IEEE. Personal use of this material is permitted. Permission from IEEE must be obtained for all other uses, in any current or future media, including reprinting/republishing this material for advertising or promotional purposes, creating new collective works, for resale or redistribution to servers or lists, or reuse of any copyrighted component of this work in other works.}}

\author{
Dada Zhang\\
  Durham School of Architectural Engineering and Construction \\
  University of Nebraska-Lincoln\\
  Lincoln, NE, 68588-0500 \\
  \texttt{dzhang19@huskers.unl.edu} \\
   \And
 Md Ruman Islam \\
  Department of Computer Science\\
  University of Nebraska-Omaha\\
  Omaha, NE, 68182 \\
  \texttt{mdrumanislam@unomaha.edu} \\
  \And
 Pei-Chi Huang \\
  Department of Computer Science\\
  University of Nebraska-Omaha\\
  Omaha, NE, 68182 \\
  \texttt{phuang@unomaha.edu} \\
  \And
 Chun-Hsing Ho \\
  Durham School of Architectural Engineering and Construction \\
  University of Nebraska-Lincoln\\
  Lincoln, NE, 68588-0500 \\
  \texttt{chunhsing.ho@unl.edu } \\
}

\begin{document}
\maketitle
\begin{abstract}
Transitioning from a Drive-by-Wire (DBW) system to a fully autonomous driving system (ADS) involves multiple stages of development and demands robust positioning and sensing capabilities. This paper presents a practice-driven framework for facilitating the DBW-to-ADS transition using a 2022 Chrysler Pacifica Hybrid Minivan equipped with cameras, LiDAR, GNSS, and onboard computing hardware configured with the Robot Operating System (ROS) and Autoware.AI. The implementation showcases offline autonomous operations utilizing pre-recorded LiDAR and camera data, point clouds, and vector maps, enabling effective localization and path planning within a structured test environment. The study addresses key challenges encountered during the transition, particularly those related to wireless-network-assisted sensing and positioning. It offers practical solutions for overcoming software incompatibility constraints, sensor synchronization issues, and limitations in real-time perception. Furthermore, the integration of sensing, data fusion, and automation is emphasized as a critical factor in supporting autonomous driving systems in map generation, simulation, and training. Overall, the transition process outlined in this work aims to provide actionable strategies for researchers pursuing DBW-to-ADS conversion. It offers direction for incorporating real-time perception, GNSS-LiDAR-camera integration, and fully ADS-equipped autonomous vehicle operations, thus contributing to the advancement of robust autonomous vehicle technologies.

\textbf{Keywords} Autonomous vehicle, DBW, Ubuntu, ROS, Autoware, Compatibility issue

\end{abstract}

\section{INTRODUCTION}
Autonomous driving refers to the capability of a vehicle to operate with partial or full autonomy without human intervention \cite{Chib}. In recent years, the development of autonomous driving systems has accelerated, driven by significant advancements in self-driving technologies. These systems aim to reduce human involvement, enhance road safety \cite{ALMASKATI2023, PETROVIC2020}, and improve transportation efficiency and convenience \cite{Parekh}. The global autonomous vehicle (AV) market has witnessed substantial growth, with approximately 31 million AVs in operation by 2019 and projections estimating an increase to 54 million by 2024 \cite{Ignatious2022}. As urban populations expand and automation technologies mature, consumer demand for vehicles equipped with autonomous features continues to rise, largely due to their potential to improve safety, reduce human error, and offer greater convenience.

Autonomous vehicles rely on a suite of sensors, including cameras, radar, LiDAR, and global positioning systems (GPS), to perceive their environment, detect objects, and respond to dynamic road conditions. However, transitioning from traditional drive-by-wire (DBW) systems, which replace mechanical linkages with electronic controls, to fully autonomous driving systems (ADS) presents significant technical challenges. While DBW systems provide a foundational architecture for automation, they were not originally designed to support the complex requirements of high-level autonomy, such as advanced perception, decision-making, and system integration.

This paper presents a practice-driven framework for transitioning DBW systems to ADS, using a Chrysler Pacifica Hybrid Minivan as a case study. The investigation includes a comprehensive review of the vehicle’s architecture, speed and steering control (SSC) capabilities, automation frameworks, and data integration strategies. A structured transition process is subsequently implemented to address the demands of autonomy. Through iterative implementation and troubleshooting efforts, particularly in resolving software incompatibility, hardware integration issues, and real-time data processing constraints. The study highlights key factors, practical constraints, and engineering trade-offs that emerge during the transition from DBW to ADS. Despite rapid progress in ADS research, a significant gap remains between simulation-driven development and practical deployment in commercial vehicles. Much of the academic literature relies on robotic platforms or simulation environments such as CARLA, which facilitate algorithm design but fail to capture the full range of system-integration and reliability challenges inherent to real-world vehicles. Concurrently, industry-driven platforms such as Apollo and Autoware provide comprehensive modular frameworks, yet their adaptation to specific vehicle architectures remains underexplored. To bridge this gap, this study introduces a systematic methodology for transitioning DBW systems to ADS through a real-vehicle case study. The proposed approach advances the practical readiness of autonomous platforms and offers empirically grounded insights into engineering trade-offs, system integration challenges, and deployment strategies. These contributions are intended to inform both researchers and practitioners, thereby supporting safer, more reliable, and scalable autonomous vehicle operations.

\section{BACKGROUND AND REVIEW OF ADS PLATFORM}
This section reviews the background of autonomous vehicle (AV) operating platforms and highlights the demand for advanced ADS through transitioning process of DBW systems. Autonomous driving (AD) relies heavily on sensors and big data analytics for accurate environmental perception, control, and decision-making (Table~\ref{tab:big_data_AV}) \cite{Ahmed2021}. Primary sensors include radar, LiDAR, cameras, inertial measurement units (IMUs), and global navigation satellite systems (GNSS) \cite{Ignatious2022}. These sensors collect continuous data that enable AVs to perceive their environment, adapt navigation paths, and interact safely with other vehicles.

\begin{table}[!t]
  \centering
  \caption{Big data generation in AVs \cite{Ahmed2021}}
  \label{tab:big_data_AV}
  \begin{tabular}{|P{2.8 cm}|P{4.3 cm}|P{7.2 cm}|}
    \hline
    \textbf{Technology} & \textbf{Components} & \textbf{Applications} \\
    \hline
    \multirow{4}{*}{\textbf{Imaging detection}} 
      & Visible or 3D cameras 
      & Blind spot, direction, side-view \\ \cline{2-3}
      & LiDAR and RADAR 
      & Recognition and detection \\ \cline{2-3}
      & Ultrasound, GPS 
      & Detection, mapping, positioning, and navigation \\
    \hline
    \multirow{3}{*}{\textbf{Cloud Computing}} 
      & Data storage and servers 
      & High-definition maps \\ \cline{2-3}
      & Communication 
      & Vehicle-to-Cloud (V2C) \\ \cline{2-3}
      & Roadside infrastructure 
      & Infotainment, vehicle sensor data \\
    \hline
    \multirow{3}{*}{\textbf{In-vehicle network}} 
      & Electronic control unit (ECU) 
      & Engine control \\ \cline{2-3}
      & Sensors, Actuators 
      & Monitoring, transmission control \\ \cline{2-3}
      & In-vehicle network 
      & Control, X-by-wire, ADAS, backbone networks \\
    \hline
  \end{tabular}
\end{table}

A key challenge in AV perception is sensor fusion, which integrates data from multiple sensors to enhance accuracy and robustness. For example, Ignatious et al. \cite{Ignatious2022} discussed the limitations of LiDAR, such as its inability to capture color information, and demonstrated how sensor fusion techniques can compensate for such deficiencies. Xiang et al. \cite{Xiang2023} conducted a comprehensive review of multi-sensor fusion methods in AD systems, identifying persistent challenges regarding computational complexity, resource demands, and hardware requirements. The limitations identified by Ignatious et al. \cite{Ignatious2022} and the challenges reviewed by Xiang et al. \cite{Xiang2023} collectively emphasize the necessity of robust, real-time data processing technologies, thereby shaping the foundation of our transition framework. To address these demands, cloud computing has been widely adopted in AV architectures for its scalable data storage and processing capabilities. However, cloud-based systems often encounter performance bottlenecks due to the high volume of data and latency constraints, making them less suitable for time-sensitive applications \cite{Arooj2022}. By contrast, edge computing decentralizes computation closer to sensors and other data sources (e.g., sensors, roadside units, and mobile devices), reducing latency and bandwidth usage, and supporting real-time decision-making \cite{Arooj2022, Biswas2023}.

\begin{figure}[t]
    \centering
    \includegraphics[width=0.5\textwidth]{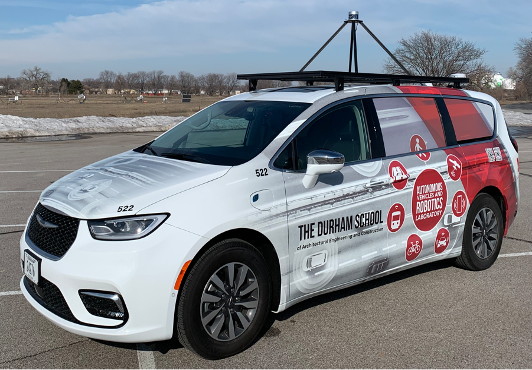}
    \caption{UNL owned autonomous vehicle.} 
    \label{fig:vehicle}
\end{figure}

AI-driven decision frameworks have become foundational to AV technology, with ongoing advancements bringing fully autonomous vehicles closer to reality \cite{Bathla2022}. However, effective deployment of these models requires access to large-scale annotated datasets. Publicly available datasets such as KITTI, nuScenes, ApolloScape, PandaSet, and Waymo \cite{Alaba2024} provide labeled sensor data and images that are essential for training supervised deep learning models for 3D object detection. We carefully reviewed these datasets to enhance AV capabilities in object recognition, localization, and tracking, contributing to the transitioning process. For practical implementation, a University of Nebraska–Lincoln (UNL)-owned Chrysler Pacifica Hybrid minivan was employed as the testing platform for developing the autonomous driving system (Figure~\ref{fig:vehicle}). The transition from a DBW configuration to a real-time autonomous driving framework required the integration of multiple core components, including data collection, processing, localization, perception, planning, control, and navigation. An initial system revealed incompatibility issues within the vehicle’s preinstalled Robot Operating System (ROS) and computing platform which needed to be addressed before advancing the transition.

Although numerous academic studies have validated autonomous systems on robotic platforms, considerable challenges remain in adapting such systems to fully autonomous commercial vehicles. To support autonomous vehicle (AV) development, both open-source and proprietary platforms are available. Open-source solutions include OpenPilot, Autoware, Apollo, and CARLA, whereas proprietary frameworks such as DriveWorks and EB Robinos provide commercial-grade alternatives \cite{Jung2025, Cha2021}. Among the open-source platforms, Apollo is notable for its deep learning-based perception and prediction modules, CARLA is widely used for simulating virtual driving environments, and Autoware offers a comprehensive self-driving software stack encompassing localization, planning, and control functionalities \cite{Cha2021, Malayjerdi2024}.

After assessing the compatibility of various ROS-based platforms with the UNL vehicle’s sensor suite including cameras, radar, GNSS, and LiDAR, we selected Autoware.AI as the primary ROS framework to support ADS development. This decision was guided by sensor integration strategies outlined by Wang et al. \cite{Wang2024} and Ren et al. \cite{Ren2023}. Consequently, the vehicle’s preinstalled ROS-compatible platform, Autoware.Auto, was upgraded to the desired version, Autoware.AI, to ensure alignment with the system requirements and enable advanced autonomous functionalities.

\section{SYSTEM SETUP: DRIVE-BY-WIRE MODE}
Following the successful installation of Autoware.AI, which replaced the previously configured Autoware.Auto, the transition process toward full autonomous vehicle operation was initiated. This section details the configuration of the drive-by-wire (DBW) system and its components, which provide the foundation for converting the UNL autonomous vehicle’s driving mechanism into an autonomous driving system. Drive-by-wire (DBW) technology—often referred to as X-by-wire—is an advanced electronic control architecture that substitutes traditional mechanical linkages with electronic actuators \cite{Goyal2019}. Since its introduction in the 1980s, DBW has emerged as a critical enabler of autonomous driving, allowing precise execution of essential vehicle functions such as throttle, steering, and braking \cite{Pillaia2022, Isermann2022}. A typical DBW system comprises an operating unit with electrical outputs that facilitate efficient and accurate control of vehicle dynamics \cite{Goyal2019}. The UNL’s autonomous vehicle used in this study is equipped with three primary DBW subsystems: steer-by-wire, throttle-by-wire, and brake-by-wire, as shown below.

\paragraph{Steer-by-wire (SBW)}
SBW consists of steering and wheel sections that use electronic signals to adjust the steering angle to enhance precision and response, particularly for adaptive driving systems and lane-keeping assist \cite{Pillaia2022}.

\paragraph{Throttle-by-wire (TBW)}
Throttle-by-wire (TBW) is also known as acceleration-by-wire. It replaces the traditional mechanical throttle linkage with an electronic throttle control unit \cite{Goyal2019}. This system reduces emissions and provides smoother acceleration, which is essential for autonomous vehicle operation \cite{Pillaia2022}.

\paragraph{Brake-by-wire (BBW)}
Brake-by-wire (BBW) is an electronically controlled braking system that replaces braking mechanisms with electronic sensors and actuators. This system reduces braking response time and enhances vehicle safety, particularly in adaptive cruise control and emergency braking systems \cite{Goyal2019, Pillaia2022}.

\begin{figure}[t]
    \centering
    \includegraphics[width=0.5\textwidth]{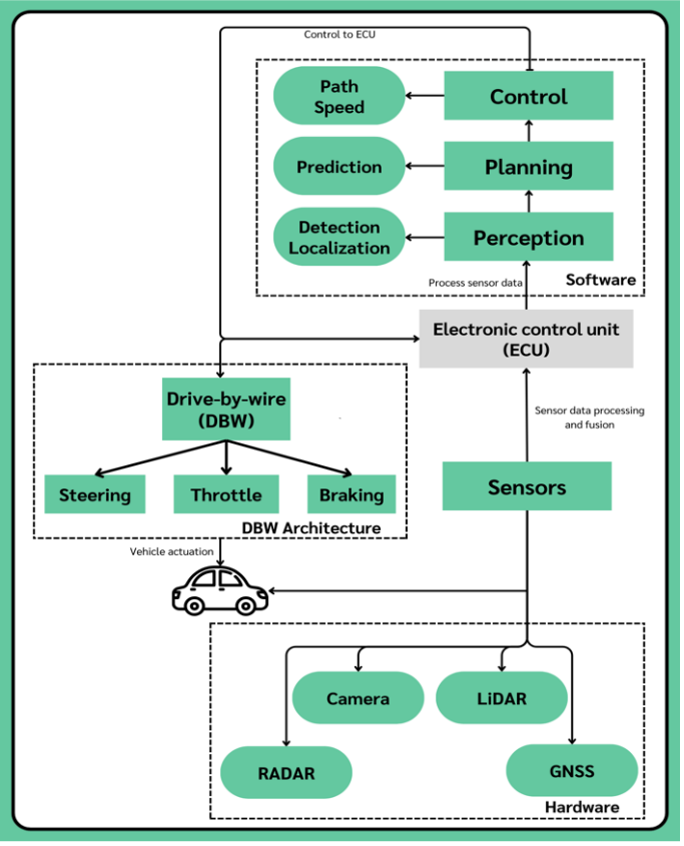}
    \caption{Architecture of autonomous driving system with DBW.} 
    \label{fig:architecture}
\end{figure}

The implementation of DBW technology in the UNL autonomous vehicle is illustrated in Figure~\ref{fig:architecture}, which depicts the interaction between Electronic Control Units (ECUs), sensors, software modules, and actuators. The operation of a fully DBW-enabled autonomous vehicle is governed by the following process.

\begin{itemize}
    \item Sensor Input: Environmental data is continuously collected via radar, LiDAR, GNSS, and cameras. The data is transmitted to the onboard computing system for perception, localization, and navigation.
    \item ECU Processing: ECU receives processed data and generates control signals based on the vehicle’s situational awareness and planned trajectory.
    \item Actuator Execution: DBW actuators interpret ECU commands and execute the corresponding vehicle maneuvers, enabling smooth, coordinated transitions between driving tasks.
\end{itemize}

\begin{figure}[t]
    \centering
    \includegraphics[width=0.6\textwidth]{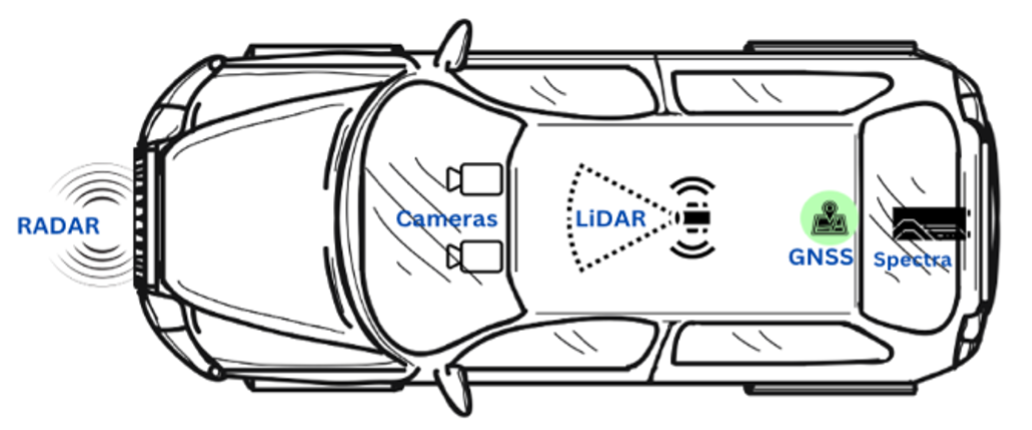}
    \caption{Sensors placement on the Chrysler Pacifica vehicle platform.} 
    \label{fig:sensors}
\end{figure}

Although DBW provides the fundamental electronic control needed for vehicles, additional layers of intelligence, such as perception, localization, and decision-making, are needed to achieve full autonomy. The following section examines the transition of the AV’s operational system from a drive-by-wire configuration to an autonomous driving mode. Implementation of Transitioning to Autonomous Driving. As shown in Figure~\ref{fig:vehicle}, a 2022 Chrysler Pacifica minivan was used as the testing platform for implementing the AD system. The vehicle is equipped with strategically mounted sensors to ensure optimal perception and data collection (Figure~\ref{fig:sensors}).

As shown in Figure~\ref{fig:sensors}, the autonomous vehicle’s radar unit is integrated into the front grille and centrally mounted to maximize object detection range and accuracy. Two Mako G-319 cameras are installed on the interior front window rack, providing stereo vision for object recognition and lane detection. A Velodyne VLP-32C LiDAR sensor is mounted on a tower structure at the front of the vehicle, offering an unobstructed forward-facing view for high-resolution 3D mapping. Additionally, the vehicle is equipped with a NovAtel GNSS receiver integrated with an internal Inertial Measurement Unit (IMU) to support precise localization.

The vehicle's software stack consists of the Robot Operating System (ROS) and an upgraded version of Autoware.AI, which offers a modular and open-source AD framework for autonomous driving. Preliminary testing confirmed that the effective operation of Autoware.AI’s packages across perception, localization, planning, and control modules, consistent with prior research \cite{Wang2024, Dhakal2021, Fan2021, autowareAIDocument}. The framework also supports sensor fusion and real-time visualization through tools such as RViz. Integration of Autoware.AI with the vehicle’s existing Ubuntu 20.04 operating system revealed incompatibility issues, primarily related to legacy dependencies on Python 2. These challenges were resolved through manual adjustments, including upgrading to Python 3, to mitigate package conflicts and ensure stable operation. Furthermore, the localization modules required GPU acceleration via NVIDIA CUDA, which introduced additional compatibility challenges. The CUDA Toolkit version was also adjusted (e.g., v11.x or v12.x for Ubuntu 20.04) based on recommendations from \cite{NVIDIA} to align with Autoware.AI requirements and enable successful deployment.

\begin{figure}[t]
    \centering
    \includegraphics[width=0.5\textwidth]{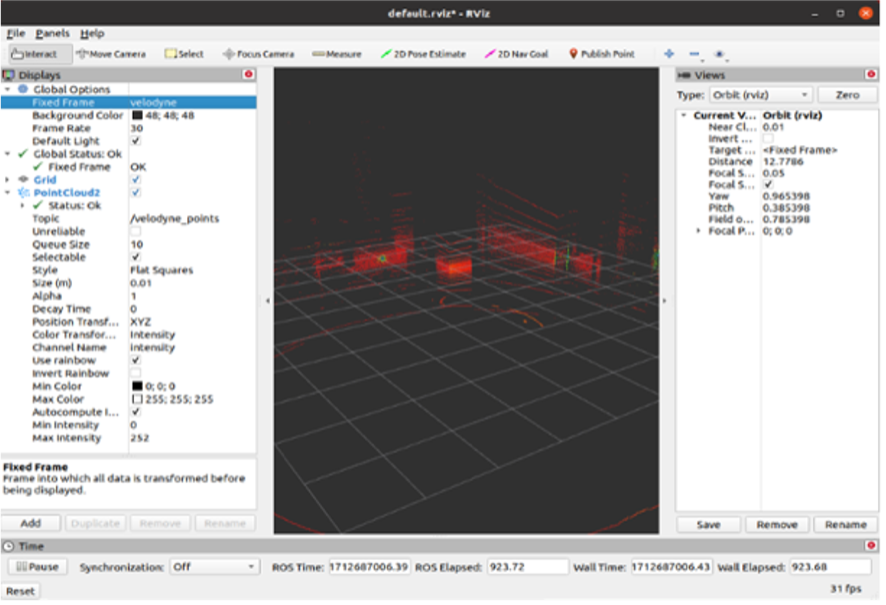}
    \caption{Visualize point cloud data in RViz using VLP-32C LiDAR.} 
    \label{fig:visualize}
\end{figure}

During the transition, sensor configuration emerged as a critical factor for ensuring real-time operation. Proper assignment and matching of IP addresses for the cameras and LiDAR within Autoware’s configuration files were required to establish stable communication. To address this, several configuration steps were implemented. Once completed, the \texttt{VLP-32C\_points.launch} file was successfully executed, enabling visualization of LiDAR data within RViz (Figure~\ref{fig:visualize}). In addition, camera parameters, such as exposure and ISO (gain), were adjusted in the camera launch file based on environmental lighting conditions. Lower exposure and ISO settings were found to be preferable in bright outdoor environments, as they enhanced image clarity and reduced overexposure.

\begin{figure}[t]
    \centering
    \includegraphics[width=0.5\textwidth]{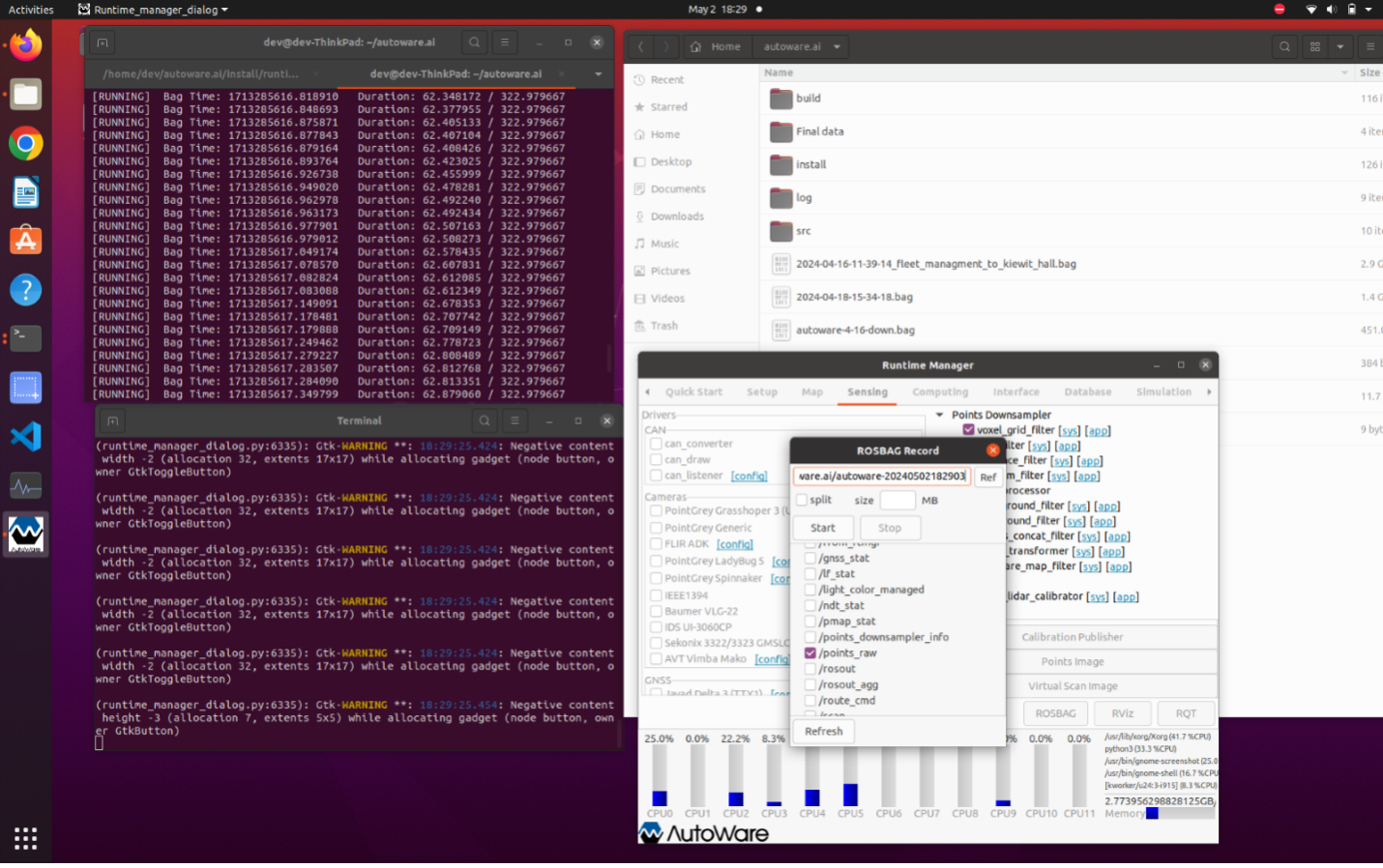}
    \caption{Downsample and process rosbag data in Autoware.} 
    \label{fig:downsample}
\end{figure}

Once all sensors were successfully integrated with the autonomous vehicle’s operating system and speed and steering control platforms, a controlled test site at the University of Nebraska–Lincoln was selected to evaluate the effectiveness of the transition implementation. The autonomous vehicle was operated along the designated test route to collect sensor data, including LiDAR and camera streams, for visualization, localization, code validation, and system analysis. Data collection was implemented using Autoware's default ROS launch files associated with the LiDAR (Velodyne) sensor. To ensure compatibility of rosbag data with RViz during playback, the appropriate reference frame (e.g., velodyne, \texttt{base\_link}) was defined during data collection. After data acquisition, downsampling was performed using the \texttt{voxel\_grid\_filter} node in Autoware.AI. This process converted high-resolution LiDAR data into more manageable formats, significantly improving processing performance. Figure~\ref{fig:downsample} presents the results of downsampling and processing of rosbag data in Autoware.AI for visualization and map generation.

\begin{figure}[t]
    \centering
    \includegraphics[width=0.5\textwidth]{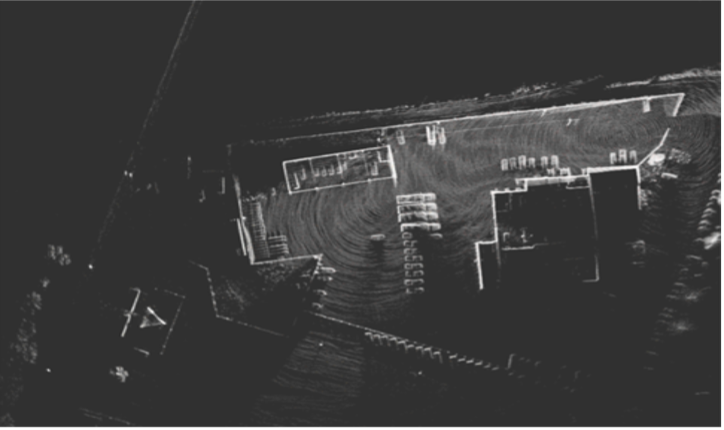}
    \caption{Visualization of PCD in RViz (test site, UNL).} 
    \label{fig:pcd}
\end{figure}

Upon completion of the downsampling process, the subsequent step in the transition toward autonomy involved generating Point Cloud Data (PCD) files using LiDAR-based localization in Autoware.AI. It is important to note that Autoware.AI requires all PCD files to be in ASCII format to ensure compatibility with its localization and mapping modules. Point cloud data initially stored in binary format can be converted using tools such as \texttt{pcl\_convert\_pcd\_ascii\_binary} from the Point Cloud Library (PCL) \cite{PCL}. This conversion ensures proper integration with Autoware.AI's mapping tools and visualization functions. Additionally, the \texttt{pcl\_viewer} tool can be employed on the PCD file. Figure~\ref{fig:pcd} illustrates the visualization of PCD data within RViz.

Additionally, a vector map is required to support path planning and high-level decision-making. Vector maps, also known as Lanelet Maps, provide semantic information, including road networks, lane geometry, and traffic lights. Two common approaches for creating vector maps include \cite{autowarecore}:

\begin{itemize}
    \item The TIER IV web-based tool (Vector Map Builder) provides a visual interface for annotating road features.
    \item Other open-source software such as MapToolbox and JOSM (Java OpenStreetMap Editor) for more advanced mapping workflows.
\end{itemize}

\begin{figure}[t]
    \centering
    \includegraphics[width=0.5\textwidth]{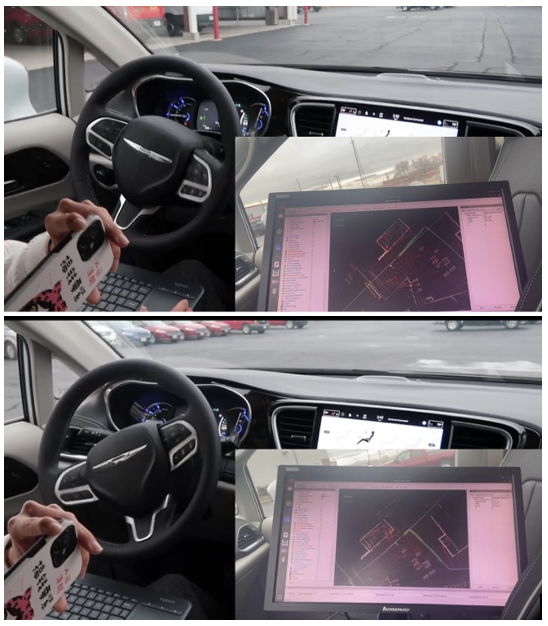}
    \caption{Validation of the autonomous driving system equipped vehicle movement and its offline operation system.} 
    \label{fig:validation}
\end{figure}

This implementation generated the vector map using TIER IV's Vector Map Builder, leveraging previously collected PCD data as the geometric reference. To validate the offline system using pre-recorded sensor data and generated maps, the vehicle's 2D pose was manually estimated using the \texttt{2D Pose Estimate} tool in RViz via Autoware.AI while the vehicle remained stationary in the testing environment. After the transition, the AV was successfully maneuvered in an autonomous mode in the controlled environment. Figure~\ref{fig:validation} illustrates the system testing process at different locations along the driving path within the vehicle.

Thus, the AD system's initial testing was conducted offline, successfully demonstrating the integration of the sensors, point cloud data, vector map, and the Autoware.AI stack. The system was able to load and visualize LiDAR and camera data in RViz and follow a predefined global path using a waypoint planner and pure pursuit in Autoware. Estimating 2D poses in RViz enabled precise vehicle positioning on the map. These results confirm the feasibility of operating the AV platform in offline autonomy mode.

\section{FUTURE WORK}
Despite significant advancements in autonomous vehicle (AV) technologies, several technical challenges persist in the transition from drive-by-wire (DBW) systems to fully autonomous driving (AD). Key areas requiring further development include sensor fusion, localization accuracy, and real-time decision-making. Based on our implementation experience, the following enhancements are proposed to advance the current system:

\begin{itemize}
    \item Integration of real-time perception using deep learning-based object detection (e.g., YOLO model) to identify dynamic and static objects in real-time.
    \item Implementation of simultaneous localization and mapping (SLAM) to enable real-time map generation and vehicle localization.
    \item End-to-end testing of real-time actuation within a fully integrated autonomous driving system.
\end{itemize}

\section{DISCUSSION}
Transitioning from DBW to a fully autonomous driving system involves multiple dimensions, including sensor configuration, perception algorithms, motion planning, and software framework integration. These components were implemented using the Robot Operating System (ROS) and Autoware.AI on a Chrysler Pacifica Hybrid equipped for autonomous operation. The successful offline deployment of autonomous functionalities established a foundation for future real-time autonomy. During the transition, we encountered several technical difficulties and developed a number of strategies to overcome the challenges. Thus, this paper offers practical “lessons learned” recommendations for researchers facing similar transition issues to undertake similar efforts.

\begin{enumerate}
\item Software version management is critical. Autoware.AI depends on specific versions of ROS and Python. Manual updates to resolve compatibility issues can delay integration and require careful planning.
\item Sensor configuration and fusion must be precise. Misalignment in network settings or parameter definitions can result in performance bottlenecks and inaccurate localization.
\item 2D pose estimation in Rviz is useful for initial training. However, if not optimized, it can slow progress and hamper autonomous performance.
\item Our experience highlights that offline autonomy is a crucial intermediate step before transitioning to real-time autonomous systems. The pre-recorded data enables extensive testing and validation of control and localization modules with minimal operational risk.
\item Continuous improvement is vital. Our goal is to continue efforts to enhance the transition from offline to real-time autonomy by integrating deep learning-based object detection into a live SLAM framework and improving obstacle avoidance capabilities.
\end{enumerate}

\section{CONCLUSIONS}
This paper presents a practice-driven approach for transitioning from a drive-by-wire system to a fully autonomous driving system, emphasizing technical implementation, system integration, and real-world operational challenges. The feasibility of this transition was demonstrated through the offline deployment of an ADS on a Chrysler Pacifica Hybrid vehicle, leveraging point cloud data and the Autoware.AI framework. Our experimental results highlight the value of this structured framework and a critical intermediate step—using pre-recorded data for extensive testing—which minimizes operational risk. This experience also yielded several key insights for researchers and practitioners undertaking similar efforts. It was found that vigilant software version management is critical, as Autoware.AI's dependencies on specific ROS and Python versions can cause significant integration delays. 

Furthermore, it was learned that precise sensor configuration and fusion are essential to prevent performance bottlenecks and ensure accurate localization. While tools like RViz are valuable for initial pose estimation, their optimization is necessary to avoid hampering autonomous performance. Future work will build upon this foundation by implementing real-time object detection using deep learning models (e.g., YOLO) and integrating a simultaneous localization and mapping (SLAM) framework. These advancements will enhance the system's responsiveness and reliability, ultimately supporting the development of safer and more efficient autonomous vehicle deployments.

\bibliographystyle{IEEEtran}
\bibliography{references}

\begin{thebibliography}{10}
\providecommand{\url}[1]{#1}
\csname url@samestyle\endcsname
\providecommand{\newblock}{\relax}
\providecommand{\bibinfo}[2]{#2}
\providecommand{\BIBentrySTDinterwordspacing}{\spaceskip=0pt\relax}
\providecommand{\BIBentryALTinterwordstretchfactor}{4}
\providecommand{\BIBentryALTinterwordspacing}{\spaceskip=\fontdimen2\font plus
\BIBentryALTinterwordstretchfactor\fontdimen3\font minus \fontdimen4\font\relax}
\providecommand{\BIBforeignlanguage}[2]{{%
\expandafter\ifx\csname l@#1\endcsname\relax
\typeout{** WARNING: IEEEtran.bst: No hyphenation pattern has been}%
\typeout{** loaded for the language `#1'. Using the pattern for}%
\typeout{** the default language instead.}%
\else
\language=\csname l@#1\endcsname
\fi
#2}}
\providecommand{\BIBdecl}{\relax}
\BIBdecl

\bibitem{Chib}
P.~S. Chib and P.~Singh, ``Recent advancements in end-to-end autonomous driving using deep learning: A survey,'' \emph{IEEE Transactions on Intelligent Vehicles}, vol.~9, no.~1, pp. 103--118, 2024.

\bibitem{ALMASKATI2023}
\BIBentryALTinterwordspacing
D.~Almaskati, S.~Kermanshachi, and A.~Pamidimukkula, ``Autonomous vehicles and traffic accidents,'' \emph{Transportation Research Procedia}, vol.~73, pp. 321--328, 2023. [Online]. Available: \url{https://www.sciencedirect.com/science/article/pii/S2352146523012231}
\BIBentrySTDinterwordspacing

\bibitem{PETROVIC2020}
\BIBentryALTinterwordspacing
Đorđe Petrović, R.~Mijailović, and D.~Pešić, ``Traffic accidents with autonomous vehicles: Type of collisions, manoeuvres and errors of conventional vehicles’ drivers,'' \emph{Transportation Research Procedia}, vol.~45, pp. 161--168, 2020. [Online]. Available: \url{https://www.sciencedirect.com/science/article/pii/S2352146520301654}
\BIBentrySTDinterwordspacing

\bibitem{Parekh}
\BIBentryALTinterwordspacing
D.~Parekh, N.~Poddar, A.~Rajpurkar, M.~Chahal, N.~Kumar, G.~P. Joshi, and W.~Cho, ``A review on autonomous vehicles: Progress, methods and challenges,'' \emph{Electronics}, vol.~11, no.~14, 2022. [Online]. Available: \url{https://www.mdpi.com/2079-9292/11/14/2162}
\BIBentrySTDinterwordspacing

\bibitem{Ignatious2022}
\BIBentryALTinterwordspacing
H.~A. Ignatious, Hesham-El-Sayed, and M.~Khan, ``An overview of sensors in autonomous vehicles,'' \emph{Procedia Computer Science}, vol. 198, pp. 736--741, 2022, 12th International Conference on Emerging Ubiquitous Systems and Pervasive Networks / 11th International Conference on Current and Future Trends of Information and Communication Technologies in Healthcare. [Online]. Available: \url{https://www.sciencedirect.com/science/article/pii/S1877050921025540}
\BIBentrySTDinterwordspacing

\bibitem{Ahmed2021}
K.~R. Ahmed, ``Big data and autonomous vehicles,'' in \emph{Deep Learning and Big Data for Intelligent Transportation}.\hskip 1em plus 0.5em minus 0.4em\relax Springer, 2021, pp. 21--47.

\bibitem{Xiang2023}
C.~Xiang, C.~Feng, X.~Xie, B.~Shi, H.~Lu, Y.~Lv, M.~Yang, and Z.~Niu, ``Multi-sensor fusion and cooperative perception for autonomous driving: A review,'' \emph{IEEE Intelligent Transportation Systems Magazine}, vol.~15, no.~5, pp. 36--58, 2023.

\bibitem{Arooj2022}
A.~Arooj, M.~S. Farooq, A.~Akram, R.~Iqbal, A.~Sharma, and G.~Dhiman, ``Big data processing and analysis in internet of vehicles: Architecture, taxonomy, and open research challenges,'' \emph{Archives of Computational Methods in Engineering}, vol.~9, no.~2, pp. 793--829, 2022.

\bibitem{Biswas2023}
\BIBentryALTinterwordspacing
A.~Biswas and H.-C. Wang, ``Autonomous vehicles enabled by the integration of {IoT}, edge intelligence, 5g, and blockchain,'' \emph{Sensors}, vol.~23, no.~4, 2023. [Online]. Available: \url{https://www.mdpi.com/1424-8220/23/4/1963}
\BIBentrySTDinterwordspacing

\bibitem{Bathla2022}
G.~Bathla, K.~Bhadane, R.~K. Singh, R.~Kumar, R.~Alvalu, R.~Krishnamurthi, A.~Kumar, R.~N. Thakur, and S.~Basheer, ``Autonomous vehicles and intelligent automation: Applications, challenges, and opportunities,'' \emph{Mobile Information Systems}, vol. 2022, no.~1, p. 7632892, 2022.

\bibitem{Alaba2024}
\BIBentryALTinterwordspacing
S.~Y. Alaba, A.~C. GurbuzORCID, and J.~E. Ball, ``Emerging trends in autonomous vehicle perception: Multimodal fusion for {3D} object detection,'' \emph{World Electric Vehicle Journal}, vol.~15, no.~1, 2024. [Online]. Available: \url{https://www.mdpi.com/2032-6653/15/1/20}
\BIBentrySTDinterwordspacing

\bibitem{Jung2025}
\BIBentryALTinterwordspacing
H.-Y. Jung, D.-H. Paek, and S.-H. Kong, ``Open-source autonomous driving software platforms: Comparison of autoware and {Apollo},'' 2025. [Online]. Available: \url{https://arxiv.org/abs/2501.18942}
\BIBentrySTDinterwordspacing

\bibitem{Cha2021}
H.~Cha and K.~Lee, ``Facilitating the development of self-driving cars with open-source projects,'' in \emph{2021 International Conference on Information and Communication Technology Convergence (ICTC)}, 2021, pp. 707--709.

\bibitem{Malayjerdi2024}
\BIBentryALTinterwordspacing
M.~Malayjerdi, R.~Sell, E.~Malayjerdi, M.~İlhan Akbaş, and R.~Razdan, ``{Real-Life} experiences in using open source for autonomy applications,'' \emph{Engineering Proceedings}, vol.~79, no.~1, 2024. [Online]. Available: \url{https://www.mdpi.com/2673-4591/79/1/19}
\BIBentrySTDinterwordspacing

\bibitem{Wang2024}
\BIBentryALTinterwordspacing
X.~Wang, M.~A. Maleki, M.~W. Azhar, and P.~Trancoso, ``Moving forward: A review of autonomous driving software and hardware systems,'' 2024. [Online]. Available: \url{https://arxiv.org/abs/2411.10291}
\BIBentrySTDinterwordspacing

\bibitem{Ren2023}
J.~Ren and D.~Xia, ``Challenges of autonomous driving systems,'' in \emph{Autonomous driving algorithms and Its IC Design}.\hskip 1em plus 0.5em minus 0.4em\relax Springer, 2023, pp. 1--23.

\bibitem{Goyal2019}
A.~Goyal and A.~Thakur, ``An overview of drive by wire technology for automobiles,'' in \emph{2019 International Conference on Automation, Computational and Technology Management (ICACTM)}, 2019, pp. 108--110.

\bibitem{Pillaia2022}
A.~V. Pillaia and M.~B, ``Overview of drive by wire technologies in automobiles,'' \emph{INTERNATIONAL CONFERENCE ON SMART GRID \& ELECTRIC VEHICLE (ICSGEV 2021)}, vol. 2452, no.~1, 2022.

\bibitem{Isermann2022}
R.~Isermann, R.~Schwarz, and S.~Stolzl, ``Fault-tolerant drive-by-wire systems,'' \emph{IEEE Control Systems Magazine}, vol.~22, no.~5, pp. 64--81, 2002.

\bibitem{Dhakal2021}
S.~Dhakal, D.~Qu, D.~Carrillo, Q.~Yang, and S.~Fu, ``{OASD}: An open approach to self-driving vehicle,'' in \emph{2021 Fourth International Conference on Connected and Autonomous Driving (MetroCAD)}, 2021, pp. 54--61.

\bibitem{Fan2021}
\BIBentryALTinterwordspacing
X.~Fan, Z.~Zhang, and Y.~Wang, ``Development of a self-driving car prototype for educational and research purposes,'' \emph{Journal of Physics: Conference Series}, vol. 1905, no.~1, p. 012013, may 2021. [Online]. Available: \url{https://doi.org/10.1088/1742-6596/1905/1/012013}
\BIBentrySTDinterwordspacing

\bibitem{autowareAIDocument}
``{Autoware.AI} documentation,'' \url{https://github.com/autowarefoundation/autoware_ai_documentation/wiki/Source-Build}, accessed: June 14, 2024.

\bibitem{NVIDIA}
``Cuda toolkit archive,'' \url{https://developer.nvidia.com/cuda-toolkit-archive}, accessed: June 14, 2024.

\bibitem{PCL}
``Point cloud library,'' \url{https://pointclouds.org/}, accessed: June 14, 2024.

\bibitem{autowarecore}
``How is autoware core/universe different from autoware.ai and autoware.auto?'' \url{https://autowarefoundation.github.io/autoware-documentation/main/design/autoware-concepts/}, 2023, accessed: June 14, 2024.

\end{thebibliography}

\end{document}